\PassOptionsToPackage{hyperref}{draft}
\documentclass[sigconf,nonacm]{acmart}
\usepackage{caption}
\usepackage{subcaption}
\usepackage{multirow}
\usepackage{xcolor}
\usepackage{graphicx}
\usepackage{epstopdf}
\usepackage{graphics}
\usepackage{url}
\usepackage{comment}
\usepackage{tabularx}
\usepackage{float}
\restylefloat{table}
\AtBeginDocument{%
  \providecommand\BibTeX{{%
  \normalfont 
  \kern-0.5em{\scshape i\kern-0.25em b}\kern-0.8em\TeX}
  }}
\graphicspath{{./llm_paper}}

\begin{document}

\title{Towards reducing hallucination in extracting information from financial reports using Large Language Models}

\author{Bhaskarjit Sarmah}
\email{bhaskarjit.sarmah@blackrock.com}
\affiliation{%
\institution{BlackRock, Inc.}
\city{Gurgaon}
\country{India}}

\author{Tianjie Zhu}
\email{tianjie.zhu@blackrock.com}
\affiliation{%
\institution{BlackRock, Inc.}
\city{Wilmington, DE}
\country{USA}}

\author{Dhagash Mehta}
\email{dhagash.mehta@blackrock.com}
\affiliation{%
\institution{BlackRock, Inc.}
\city{New York, NY}
\country{USA}}

\author{Stefano Pasquali}
\email{stefano.pasquali@blackrock.com}
\affiliation{%
\institution{BlackRock, Inc.}
\city{New York, NY}
\country{USA}}

\renewcommand{\shortauthors}{Sarmah et al.}

\begin{abstract}
For a financial analyst, the question and answer (Q\&A) segment of the company financial report is a crucial piece of information for various analysis and investment decisions. However, extracting valuable insights from the Q\&A section has posed considerable challenges as the conventional methods such as detailed reading and note-taking lack scalability and are susceptible to human errors, and Optical Character Recognition (OCR) and similar techniques encounter difficulties in accurately processing unstructured transcript text, often missing subtle linguistic nuances that drive investor decisions. Here, we demonstrate the utilization of Large Language Models (LLMs) to efficiently and rapidly extract information from earnings report transcripts while ensuring high accuracy—transforming the extraction process as well as reducing hallucination by combining retrieval-augmented generation technique as well as metadata. We evaluate the outcomes of various LLMs with and without using our proposed approach based on various objective metrics for evaluating Q\&A systems, and empirically demonstrate superiority of our method.
\end{abstract}




\keywords{Large Language Models, Financial Markets, Earning Call Transcripts, Natural Language Processing}

\maketitle

\section{Introduction}
The Q\&A section of corporate earnings reports serves as a valuable repository of insights that guide investors, financial analysts, and stakeholders in making informed decisions. However, researchers face the arduous task of manually parsing through voluminous reports to identify pertinent questions and answers. The reliance on traditional methods, such as keyword matching and rule-based approaches, often results in limited accuracy and scalability.

Previous studies such as Sheikh et al.~\cite{sheikh2012rule} (see also Refs.~\cite{qian2018graphie,patel2020abstractive,liu2019graph}) attempted to automate this process using rule-based techniques like Greedy and Tabu Search algorithms, but their effectiveness was constrained by the complexity and diversity of linguistic structures present in these reports. Although these approaches achieve high recall, they lack precision compared to human analysts. In general, the intricate nature of financial terminology and domain-specific language presents obstacles for rule-based approaches which struggle to adapt to earnings reports' dynamic language patterns.

Optical Character Recognition (OCR) based techniques which attempts to automate information extraction rely on advanced image processing and natural language understanding technologies as mentioned in Refs.~\cite{ha2022information,cutting2021intelligent}. Initially, OCR algorithms preprocess scanned or digital images of financial reports segmenting text regions from graphical elements. Then, character recognition engines analyze these segments and convert them into machine-readable text. To identify key financial data points, rule-based or ML models parse the recognized text, locating relevant information like revenue, expenses, and net income. Natural Language Processing (NLP) techniques may contextualize textual content, aiding in trend analysis and sentiment extraction. However, the OCR approach usually does not have context of the information within the document and unable understand the document for abstractive Q\&A system if not paired with other text-based approach \cite{vsimsa2023docile}

With the emergence of LLMs such as BERT \cite{devlin2018bert} and GPT-3 \cite{brown2020language}, the landscape of information extraction from earnings reports has been revolutionized. These models leverage their inherent understanding of contextual nuances to accurately identify and extract relevant question-answer pairs. Researchers now benefit from a data-driven approach that adapts to the dynamic language patterns in earnings reports, thereby enhancing both efficiency and precision. 

Here, we present a comparative analysis of various pre-trained LLMs having capabilities such as grasping contextual cues and syntactic structures. Furthermore, we highlight recent studies such as Refs.~\cite{liu2023reta,yue2023leveraging,ni2023paradigm} that demonstrate the efficacy of LLMs in surpassing previous limitations and achieving state-of-the-art results in other domains. In short, the integration of LLMs into the extraction of question-answer sections from corporate earnings reports has revolutionized the decision-making landscape for financial analysts.

However, these systems frequently exhibit tendencies toward generating responses that deviate from factual accuracy, often termed as \textit{hallucination}, as observed in Ref.~\cite{zhuang2023toolqa}. Addressing this critical issue, researchers have proactively introduced an innovative remedy by enhancing LLMs through the seamless integration of retrieval systems, as demonstrated by Ref.~\cite{lewis2020retrieval}. These retrieval-augmented LLMs aim to bolster the system's accuracy by harnessing external repositories of information. However, even with these advancements, challenges persist, particularly in scenarios involving the processing of multiple documents. In such cases, the model might inadvertently extract information from unintended sources, especially when questions bear resemblance. This unintended extraction raises concerns about the emergence of hallucinatory responses, where answers are derived from entities not directly related to the query. When we build a Q\&A system with multiple documents at once, there may be questions which are related to all the documents in the database. Hence, it is indeed possible that while answering to generalized questions, an LLM may fetch results from multiple other documents which are not related to the document in question and, in turn, may provide incorrect answers.

We take a comprehensive approach to address this multifaceted challenge, by not only integrating retrieval-augmented LLMs but also leveraging metadata to effectively mitigate the occurrence of such hallucinatory responses. Through this endeavor, we contribute to enhancing the reliability and precision of information extraction from the LLMs, ensuring that responses align more closely with the actual context and requirements of the queries posed by users.

\section{Data}
We used transcripts from earnings calls of Nifty 50 constituents for our analysis. This dataset is widely recognized in the investment realm and is esteemed as an authoritative and extensive collection of earnings call transcripts. In our investigation, we focus on data spanning the quarter ending in June, 2023 i.e. the earnings reports for Q1 of the financial year 2024\footnote{An Indian financial year for 2024 starts on 1st April 2023 and ends in 31st March 2024, so the quarter from 1st April to 30th June is Q1 of financial year 2024.}. Our dataset encompasses 50 transcripts for this quarter, spanning over 50 companies within Nifty 50 universe from different sectors including Infrastructure, Healthcare, Consumer Durables, Banking, Automobile, Financial Services, Energy - Oil \& Gas, Telecommunication, Consumer Goods, Pharmaceuticals, Energy - Coal, 
Materials, Information Technology, Construction, Diversified, Metals, Energy - Power and Chemicals providing a substantial and diverse foundation for our study.

\section{Methodology}
While LLMs have proven to be incredibly powerful tools for various natural language processing tasks, they do come with certain limitations, particularly when it comes to extracting information from a recent time period: LLMs are trained on vast amounts of data up until a certain cutoff point in time, after which they lack access to new information or context that has emerged post-training. This means that when attempting to extract information or insights from a time period beyond their training data, LLMs might be ill-equipped to provide accurate and up-to-date responses. Since their knowledge is essentially fixed at their last update, any developments or events occurring after the cutoff time are beyond their scope. This limitation underscores the importance of understanding the temporal boundaries of LLMs' knowledge and employing alternative methods, such as real-time data sources or more recent model updates, when seeking information from contemporary contexts.

Retrieval-augmented generation \cite{lewis2020retrieval} represents a paradigm shift in the landscape of large language models (LLMs) by addressing one of their key limitations—hallucinations and inaccuracies. This innovative approach enhances the capabilities of LLMs by integrating retrieval systems into their architecture. Traditionally, LLMs solely generate responses based on their learned patterns from training data, often resulting in the production of plausible yet factually incorrect information. The retrieval-augmented LLMs, on the other hand, introduce an external knowledge source—an information repository—from which they retrieve relevant content to inform their generated responses. This process enables LLMs to ground their outputs in verifiable and contextually relevant information, reducing the propensity for generating false or misleading content. By combining the contextual understanding and creativity of LLMs with the precision of retrieval systems, this approach offers a harmonious synergy that elevates the quality of generated text.

However, in scenarios involving the processing of multiple documents, there is a possibility that the model could retrieve information from unrelated entities when faced with identical questions. This situation could lead to a errors while extracting information where the model generates answers from incorrect sources. To address this challenge, our study aims to mitigate such issues by leveraging metadata assistance.

\subsection{Data Collection}
We start the data collection process focused on acquiring earnings reports from company websites within the Nifty 50 universe. To achieve this, a custom web scraping tool was developed and deployed. This tool was designed to navigate through the websites of each company within the Nifty 50 index, systematically retrieving the pertinent earnings reports for Q1 of the financial year 2024. By utilizing this web scraping approach, we aimed to compile a comprehensive dataset encompassing the earnings reports of the constituent companies. This dataset forms a critical foundation for our subsequent analysis and investigation into information extraction using large language models.

Table \ref{tab:summary} summarizes basic statistics of the documents we will be experimenting with in the remainder of this work.

\begin{table}[ht]
\centering
\begin{tabular}{|l|c|}
\hline
Number of companies/documents & 50 \\
Average number of pages & 27 \\
Average number of questions & 16 \\
Average number of tokens & 60,000 \\
\hline
\end{tabular}
\caption{Summary Statistics for the call transcript documents used in the present work.}
\label{tab:summary}
\end{table}
\vspace{-0.25in}
\subsection{LLM}
For this experiment we used text-bison model\footnote{https://cloud.google.com/vertex-ai/docs/generative-ai/model-reference/text} available in Google Cloud Platoform. The text-bison model is a foundation model for natural language processing tasks on Google Cloud Platform (GCP). It is optimized for a variety of tasks, including sentiment analysis, entity extraction, and content creation. The model is trained on a massive dataset of text and code, and it can be used to generate text, translate languages, write different kinds of creative content, and answer your questions in an informative way. Its maximum input token capacity is 8192, while the output token capacity is 1024, and the training data encompasses information up until February 2023. 

In the present study, we experiment with multiple open-source pretrained LLMs which are listed along with their architectures and maximum token size in Table \ref{tab:models}.
\begin{table}[ht]
\centering
\begin{tabularx}{\linewidth}{|c|c|c|c|}
\hline
Model & Maintainer & Architecture & Token \\
\hline
PaLM2 \cite{anil2023palm} & Google & Encoder - Decoder & 8192 \\
Bloomz-7b1 \cite{muennighoff2022crosslingual} & BigScience & Decoder & 2048 \\
Pythia-1.4b \cite{biderman2023pythia} & EleutherAI & Decoder & 2048 \\
flan-t5-base \cite{chung2022scaling} & Google & Encoder - Decoder & 512 \\
Llama2-7b \cite{touvron2023llama} & Meta & Decoder & 4096 \\
\hline
\end{tabularx}
\caption{Summary of LLMs used in the present work.}
\label{tab:models}
\end{table}

We also used below parameters across all the model configuration settings to remain consistent in the experiments. 
\begin{table}[ht]
\centering
\begin{tabular}{|l|c|}
\hline
\textbf{Parameter} & \textbf{Value} \\
\hline
Chunk Size & 1024 \\
Chunk Overlap & 0 \\
Maximum Output Tokens & 1024 \\
Chunks for Similarity Algorithm & 20 \\
Chunks to Add to Prompt & 4 \\
\hline
\end{tabular}
\caption{Model Configuration Parameters}
\label{tab:parameters}
\end{table}

\begin{figure}[h]
\includegraphics[width=8cm]{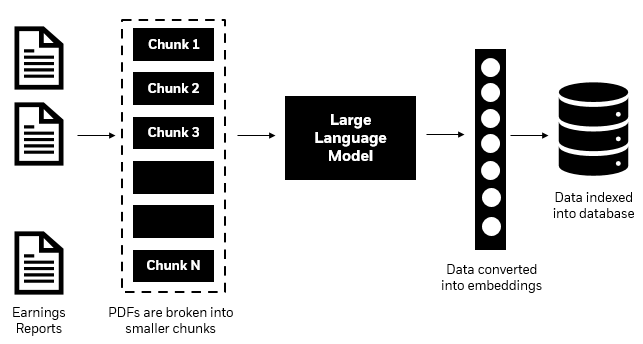}
\caption{A schematic diagram describing the chunking process for LLMs with limited token words.}
\label{fig:chunking}
\end{figure}


\subsection{Data Splitting}
As discussed, LLMs may grapple with a specific constraint of "context window" which delineates the boundaries within which these models effectively process text. For example, the text-bison model has context window of 8192 tokens which is equivalent to 7000 words approximately and output token of 1024 tokens, roughly equating to 875 words.

However, what happens when these models are provided with a document surpassing their context window? This is where a smart approach termed "chunking" comes into action. Chunking entails breaking down the document into smaller, more manageable segments that comfortably fit within the context window of the expansive language model, as shown in Figure \ref{fig:chunking}.

In this work, we deploy \emph{RecursiveCharacterTextSplitter}, which operates by segmenting extensive text into chunks of a designated size, facilitated by a defined set of characters. The default character set ["\textbackslash n \textbackslash n", "\textbackslash n", " ", ""] is utilized for this purpose. When applied to the given large text, the \emph{RecursiveCharacterTextSplitter} initially attempts to split it based on the occurrence of the first character sequence "\textbackslash n \textbackslash n." If this initial split still results in a segment larger than the specified chunk size, the process continues to the next character, "\textbackslash n," for further splitting. This sequence of attempts persists until a split is achieved that meets the criteria of being smaller than the specified chunk size.

\begin{table*}[h!]
    \centering
    \begin{tabular}{lccp{2cm}p{2cm}p{2cm}p{2cm}}
        \hline
        & & BERTScore & BARTScore & Jaro Similarity & LCSubsequence Similarity & LCSubsequence Word Count \\
        \hline
        \multicolumn{1}{l}{PaLM2} & Without Metadata & 0.59 & -3.45 & 0.66 & 0.37 & 134 \\
        & With Metadata & \textbf{0.60} & \textbf{-3.41} & \textbf{0.66} & \textbf{0.37} & \textbf{201} \\
        \hline
        \multicolumn{1}{l}{BLOOMZ-7b1} & Without Metadata & 0.35 & -4.44 & 0.54 & 0.07 & \textbf{130} \\
        & With Metadata & \textbf{0.36} & \textbf{-4.43} & \textbf{0.54} & \textbf{0.07} & 16 \\
        \hline
        \multicolumn{1}{l}{Pythia-1.4b-deduped} & Without Metadata & 0.48 & \textbf{-2.75} & \textbf{0.53} & \textbf{0.09} & 134 \\
        & With Metadata & \textbf{0.50} & -3.00 & \textbf{0.53} & \textbf{0.09} & \textbf{267} \\
        \hline
        \multicolumn{1}{l}{flan-t5-base} & Without Metadata & 0.44 & \textbf{-4.18} & 0.58 & 0.20 & 68 \\
        & With Metadata & \textbf{0.46} & -3.72 & \textbf{0.59} & \textbf{0.24} & \textbf{90} \\
        \hline
        \multicolumn{1}{l}{Llama2-7b} & Without Metadata & \textbf{0.54} & \textbf{-3.55} & \textbf{0.61} & \textbf{0.26} & 127 \\
        & With Metadata & 0.53 & \textbf{-3.43} & 0.61 & 0.25 & \textbf{171} \\
        \hline
    \end{tabular}
    \caption{Comparison of different evaluation metrics for with and without metadata for different pretrained LLMs.}\label{tab:LLM_scores}
\end{table*}

\subsection{Generating the Embeddings}
Upon segmenting the earnings reports into discrete units of predetermined dimensions, we proceed to derive embeddings for these sections through the utilization of the LLMs. These embeddings are subsequently archived within a designated database, which consequently serves as a retriever for the development of a Q\&A system specific to the earnings documents. When a user submits a query, the corresponding query vector is also transformed into an embedding. And then we use Maximal Marginal Relevance (MMR) \cite{goldstein1998summarization} for retrieving answers based on
query relevance and information novelty. The MMR criterion is used to balance the reduction of redundancy while maintaining query relevance when re-ranking retrieved documents and selecting documents for retrieving.
However, this approach carries the potential of yielding inaccurate outputs on occasion. This is due to the diverse range of documents spanning multiple companies, where identical questions could hold relevance across all these companies. To illustrate, consider the following examples of questions that pertain to all companies:

\begin{enumerate}
  \item What inquiries did analysts raise in this quarter?
  \item What pertinent details exist regarding industry trends for the quarter?
  \item Provide any insights into mergers and acquisitions (M\&A) deliberated during the call.
  \item What essential and recurring subjects were discussed during the call?
  \item What is the holistic outlook for the business?
\end{enumerate}

To address this challenge, it becomes imperative to leverage metadata. In the process of responding to these questions, we must specify the company documents we intend to examine. We employ document metadata as a filter to refine the document selection and extract relevant answers tailored to the user's query. Metadata plays a pivotal role in this context by offering crucial information about the documents, such as their source, industry, date, and more. This metadata is meticulously curated and stored, providing an organized structure that aids in efficient retrieval and filtering of information. 

The present work elucidates the methodology behind metadata creation, its storage mechanisms, and its integral role in enhancing the accuracy and precision of the information extraction process.
The utilization of metadata during document retrieval ensures a targeted approach: by considering the company associated with the question, this strategy narrows down the search to documents pertinent to that specific company, consequently mitigating the risk of obtaining inaccurate answers.

\subsection{Evaluation Methodology}
To evaluate the quality of the answers generated by the respective LLMs, we compare the answer to each of the questions to their ground truth answers that are manually created for total 400 questions as below:

\subsubsection{Ground-truth Labels}
We meticulously examined the earnings reports within the Nifty50 universe, systematically curated a comprehensive array of randomly selected 400 questions posed during the earnings calls from all the documents, and gathered the exact verbatim responses corresponding to these questions. These questions constituted the specific queries articulated by financial analysts to the management during these calls.

\subsubsection{Evaluation Metrics}
Given the ground-truth answers to 400 questions, we compute the following scores for the answers generated by each of the LLMs:

\textbf{BERTScore}\cite{Zhang*2020BERTScore} is a measure of semantic similarity that employs contextualized embeddings derived from pre-trained BERT model to estimate the semantic closeness between two pieces of text.

\textbf{BARTScore}\cite{NEURIPS2021_e4d2b6e6} is a measure of semantic similarity that utilizes the pre-trained BART model and that achieves high correlation with human judgements.

\textbf{Jaro similarity} \cite{jaro1989advances,winkler1990string} is a measure of similarity based on matching symbols and transpositions in two strings.

\textbf{Longest Common (LC) Subsequence} \cite{bergroth2000survey} similarity is a degree of similarity between two strings based on the length of their longest common subsequence. We also use the word count variant of LC Subsequence in this work.

\section{Results}
Our results are summarized in Table \ref{tab:LLM_scores}. In the table, for each of the LLMs from Table \ref{tab:models}, the final evaluation metric scores are recorded for two cases: with using metadata, and without using metadata.

All the scores mentioned above are the higher the better. The final scores mentioned in Table \ref{tab:LLM_scores} are averaged over the 400 question-answer pairs. The results suggest that using metadata, most of the LLMs yield answers closer to the ground truth answers that are curated from the Q\&A section of the earning reports, as opposed to the same LLMs without using metadata. This conclusion is most pronounced for the LCSubsequence Word Count and BertScore metrics. With respect to the remaining metrics though, the difference between the two approaches appears to be less pronounced, possibly due to the utilization of embeddings from an external model as the foundation for string matching. This aspect warrants additional investigation and scrutiny.

Moreover, when evaluating semantic similarity-based metrics, the PaLM2 model outperforms all others for BERT Score while Pythia performs best in BART Score, whereas in the context of pairwise alignment-based similarity metrics, the PaLM2 model exhibits superior performance when compared to its counterparts.
\vspace{-0.16in}
\section{Conclusions}
In this work, we employed data retrieval process involving the collection of earnings call reports within the Nifty50 universe and addressed a crucial challenge related to the contextual limitations of the models utilized: we first divided these PDF files into multiple segments, aligning with the specific context window of the models in use. These segmented PDF documents, along with their associated metadata, were then organized and stored in a vector database. When responding to user queries about a particular company, we leveraged the metadata information to selectively filter and retrieve only those document segments pertinent to the queried company. This strategic filtering significantly mitigated model hallucination issues, particularly within the framework of our multi-document Q\&A system. Moreover, we employed a maximal marginal relevance similarity search technique exclusively on the filtered document segments within the vector database.

This search allowed us to identify semantically similar document chunks, which were subsequently integrated into a coherent question-answering chain. This approach enhanced the accuracy and relevance of our generated answers as measured by various evaluation metrics and with respect to the manually curated, ground-truth, answers to 400 questions. To the best of our knowledge, this is one of the first such studies of evaluation metrics for Q\&A system for financial documents employing LLMs.

Several key strategies warrant further exploration such as use of dimensional reduction techniques to mitigate the curse of dimensionality; adjusting the number of retrieved documents that are incorporated into the context for generating answers holds and fine-tuning based on this aspect; etc.

\section*{Acknowledgement}
The views expressed here are those of the authors alone and not of BlackRock, Inc.

\bibliographystyle{ACM-Reference-Format}
\bibliography{main}

\end{document}